\documentclass[preprint,12pt,authoryear]{elsarticle}

\usepackage{amssymb}
\usepackage{amsmath}
\usepackage{graphicx}
\usepackage{booktabs} 
\usepackage{multirow} 
\usepackage{geometry} 
\usepackage{hyperref} 
\usepackage{caption}
\usepackage{subcaption}
\usepackage{float}
\usepackage{amssymb}
\usepackage{lineno}
\usepackage{xspace}
\usepackage{color}
\usepackage{comment}

\newcommand{\ourmodel}{SPIN\xspace}
\newcommand{\PMcoarse}{$\mathrm{PM}_{10}$\xspace}
\newcommand{\PM}{$\mathrm{PM}_{2.5}$\xspace}

\newcommand{\ug}{$\mu\mathrm{g}/\mathrm{m}^3$\xspace}
\newcommand{\Ammonia}{$\mathrm{NH}_3$\xspace}
\newcommand{\NitrogenOxides}{$\mathrm{NO}_x$\xspace}
\newcommand{\SulfurDioxide}{$\mathrm{SO}_2$\xspace}
\newcommand{\VOC}{$\mathrm{VOC}$\xspace}

\journal{Sustainable Cities and Society}

\begin{document}


\begin{frontmatter}


\title{Physics-Guided Inductive Spatiotemporal Kriging for PM$_{2.5}$ with Satellite Gradient Constraints}

\makeatletter
\def\thefootnote{\fnsymbol{footnote}}
\makeatother

\makeatletter
\def\thefootnote{\fnsymbol{footnote}}
\makeatother

\author[1,3]{Shuo Wang}
\ead{shuowang.ai@gmail.com}

\author[2]{Mengfan Teng\corref{cor1}}
\ead{tengmengfan@whu.edu.cn}

\author[3]{Yun Cheng}
\author[3]{Lothar Thiele}
\author[4]{Olga Saukh}
\author[5]{Shuangshuang He}
\author[5]{Yuanting Zhang}
\author[1]{Jiang Zhang\corref{cor1}}
\ead{zhangjiang@bnu.edu.cn}
\author[6]{Gangfeng Zhang}
\author[5]{Xingyuan Yuan}
\author[1]{Jingfang Fan}

\cortext[cor1]{Corresponding authors.}

\address[1]{School of Systems Science, Beijing Normal University, Beijing, China}
\address[2]{College of Information Engineering, Jiangxi University of Science and Technology, Ganzhou, China}
\address[3]{D-ITET, ETH Zurich, Zurich, Switzerland}
\address[4]{Graz University of Technology, Graz, Austria}
\address[5]{ColorfulClouds Technology Co.,Ltd., Beijing, China}
\address[6]{State Key Laboratory of Earth Surface Processes and Disaster Risk Reduction, Beijing Normal University, Beijing, China}

\begin{abstract}
High-resolution mapping of fine particulate matter (\PM) is a cornerstone of sustainable urbanism but remains critically hindered by the spatial sparsity of ground monitoring networks. While traditional data-driven methods attempt to bridge this gap using satellite Aerosol Optical Depth (AOD), they often suffer from severe, non-random data missingness (e.g., due to cloud cover or nighttime) and inversion biases. To overcome these limitations, this study proposes the \textbf{S}patiotemporal \textbf{P}hysics-Guided \textbf{I}nference \textbf{N}etwork (\ourmodel), a novel framework designed for inductive spatiotemporal kriging. Unlike conventional approaches, \ourmodel\ synergistically integrates domain knowledge into deep learning by explicitly modeling physical advection and diffusion processes via parallel graph kernels. Crucially, we introduce a paradigm-shifting training strategy: rather than using error-prone AOD as a direct input, we repurpose it as a \textit{spatial gradient constraint} within the loss function. This allows the model to learn structural pollution patterns from satellite data while remaining robust to data voids. Validated in the highly polluted Beijing-Tianjin-Hebei and Surrounding Areas (BTHSA), \ourmodel\ achieves a new state-of-the-art with a Mean Absolute Error (MAE) of 9.52 \ug, effectively generating continuous, physically plausible pollution fields even in unmonitored areas. This work provides a robust, low-cost, and all-weather solution for fine-grained environmental management.
\end{abstract}




\end{frontmatter}



\section{Introduction}
\label{sec:introduction}

\subsection{From Forecasting to Inference: The Spatial Gap in Air Quality Intelligence}
Fine particulate matter (\PM) remains a formidable threat to global public health and ecological stability~\citep{keller2015unified, burnett2018global}. Long-term exposure to elevated \PM levels is intrinsically linked to cardiovascular~\citep{chen2017detecting, xi2022effects} and respiratory diseases~\citep{chai2019effect}, necessitating precise monitoring for effective policymaking. Recently, breakthrough efforts in gapless global and regional mapping~\citep{geng2021tracking, southerland2022global, wei2023gapless} have revealed a critical discrepancy: sparsely distributed ground stations frequently fail to capture fine-grained pollution hotspots, thereby fundamentally limiting the accuracy of community-level exposure assessments.

While the past decade has witnessed remarkable progress in \textit{spatiotemporal forecasting at fixed monitoring sites}, driven by advanced deep learning models~\citep{qi2019hybrid, wang2020pm2, xiao2022dual, ma2025causal, wang2025pcdcnet, teng202372, teng2024new}, a pervasive challenge persists: the \textit{spatial data gap}. Although these approaches have achieved high precision in predicting future states, they are predominantly \textbf{transductive} in nature—fundamentally relying on historical local data streams and fixed graph topologies of existing stations. This dependency renders them ineffective for the vast unmonitored "blind spots" where no historical ground truth exists. Consequently, to achieve holistic environmental intelligence, the research frontier must expand from \textbf{fixed-point forecasting} to \textbf{inductive spatiotemporal kriging}—the capability to reconstruct continuous, high-resolution pollution fields for arbitrary unobserved locations based on sparse point measurements.

\subsection{Challenges in Current Modeling Paradigms}
Current approaches to high-resolution mapping predominantly fall into two categories, each with distinct limitations that hinder real-time, all-weather application.

\textbf{Chemical Transport Models (CTMs):} Physics-based models, such as CMAQ and WRF-Chem, represent the traditional gold standard, offering interpretability by numerically solving atmospheric dynamic equations. However, their operational utility is often constrained by prohibitive computational costs for real-time, high-resolution simulations and a heavy reliance on emission inventories that often lag behind real-time changes~\citep{woody2016multiscale, wang2025pcdcnet}. As noted in \citet{wang2025pcdcnet}, while surrogate models are emerging to address computational bottlenecks, explicitly integrating physical constraints into data-driven inference remains a grand challenge in Earth system science~\citep{reichstein2019deep}.

\textbf{Data-Driven Deep Learning \& AOD Utilization:} In the data-driven domain, Graph Neural Networks (GNNs) have emerged as the dominant paradigm for capturing complex spatiotemporal dependencies~\citep{li2018diffusion, wang2020pm2, xiao2022dual, ma2025causal}. However, a critical limitation of most existing architectures is their \textit{transductive} nature—they typically rely on fixed graph structures defined during training and lack the inductive capability to generalize to unobserved locations outside the training graph. 

To overcome station sparsity, many studies incorporate satellite Aerosol Optical Depth (AOD) as an auxiliary spatial input~\citep{van2016global, wei2021himawari, teng202372}. Pioneering works like \citet{wei2021reconstructing} and \citet{xiao2017full} have successfully utilized AOD for high-quality historical reconstruction. However, relying on AOD as a \textit{direct input feature} presents a fundamental operational dilemma: satellite retrievals suffer from severe, non-random missingness due to cloud cover and high surface reflectance (e.g., snow/ice)~\citep{levy2013collection, zhang2019validation}. Consequently, models treating AOD as a mandatory input often fail or produce significant artifacts in these data-void regions, severely limiting their robustness for real-time, all-weather operational systems.

\subsection{A Physics-Guided Inductive Framework}
To bridge these gaps and answer the call for Physics-Informed Machine Learning (PIML)~\citep{karniadakis2021physics}, this study proposes the \textbf{S}patiotemporal \textbf{P}hysics-Guided \textbf{I}nference \textbf{N}etwork (\ourmodel). \ourmodel\ distinguishes itself by embedding physical laws into the deep learning architecture and treating satellite data as a constraint rather than an input. The primary contributions are fourfold:

\begin{enumerate}
    \item \textbf{Physics-Embedded Architecture:} We propose a novel graph network architecture that explicitly embeds fundamental atmospheric processes. By designing parallel graph kernels for advection (wind-driven transport) and diffusion, the model learns spatiotemporal dependencies consistent with physical laws, surpassing the "black-box" nature of traditional deep learning.
    
    \item \textbf{Inductive Inference Mechanism:} The framework is explicitly designed for inductive kriging. Through a dynamic training strategy that randomly partitions observed and unobserved nodes, \ourmodel\ learns a generalizable interpolation function. This enables it to accurately predict \PM concentrations at any arbitrary unobserved location (\textit{Station Inference}) and generate complete maps (\textit{Grid Inference}), effectively extending prediction capabilities from discrete sites to the global space.
    
    \item \textbf{Paradigm-Shifting AOD Fusion:} We introduce a novel \textit{Masked AOD Spatial Gradient Loss}. Instead of using AOD as a direct, error-prone input feature (which fails under cloud cover), we use the \textit{spatial gradients} of available AOD data to constrain the model's output structure during training. This approach robustly leverages the structural patterns in satellite imagery while elegantly circumventing the pervasive issue of missing data.
    
    \item \textbf{SOTA Performance:} Through extensive experiments in the highly polluted Beijing-Tianjin-Hebei and Surrounding Areas (BTHSA), demonstrate that \ourmodel\ significantly outperforms comprehensive baselines—spanning feature-based, temporal, and state-of-the-art spatiotemporal kriging methods—achieving a new benchmark with a Mean Absolute Error (MAE) of 9.52 \ug.
\end{enumerate}

\section{Materials and Methods}
\label{sec:methods}

\subsection{Study Domain and Multi-Source Datasets}
\begin{figure}[H]
    \centering
    \includegraphics[width=0.95\textwidth]{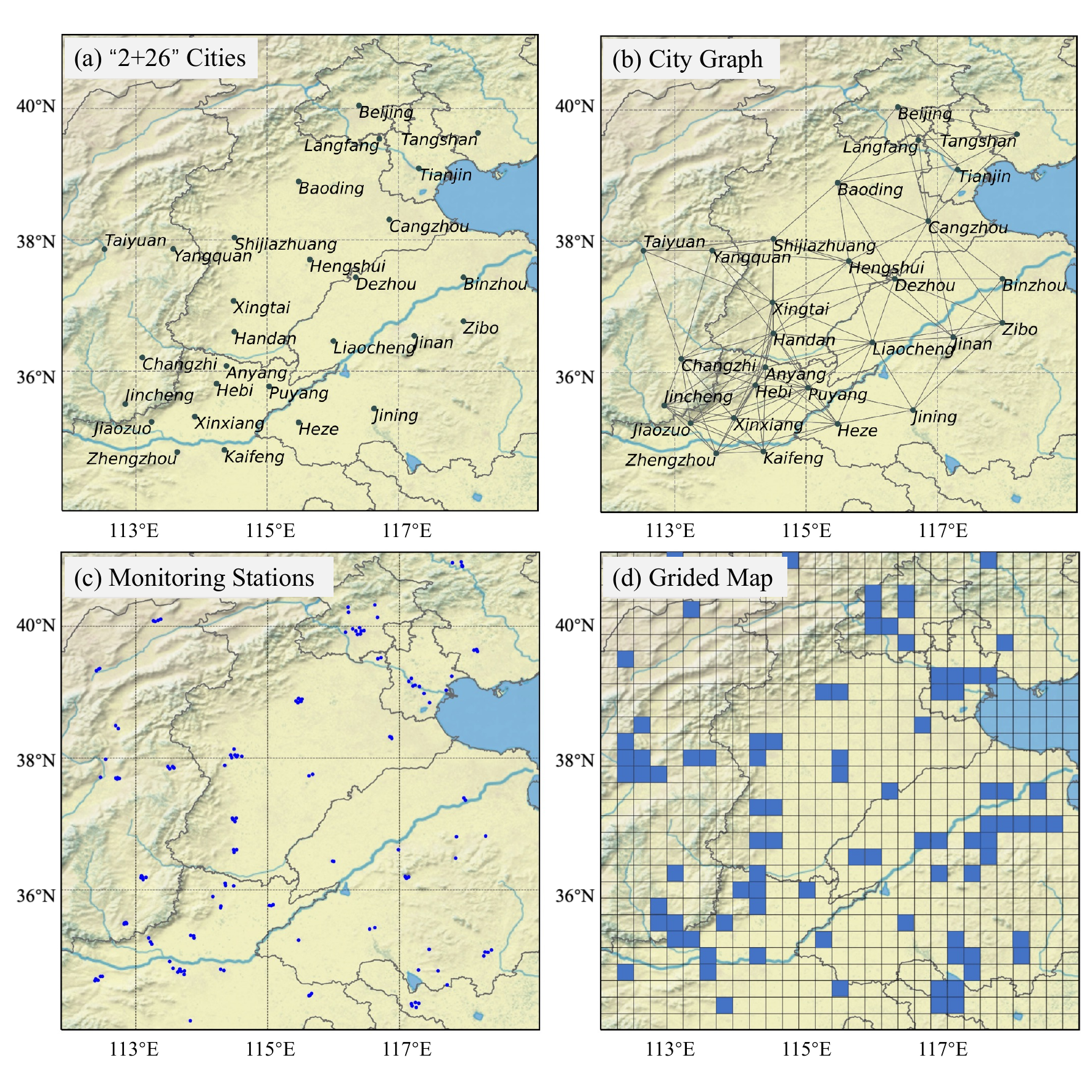}
    \caption{\textbf{Study area and task setup.} (a) The "2+26" cities in the BTHSA region. (b) A spatial adjacency graph constructed by connecting monitoring stations within a \textbf{200km geodesic threshold}~\citep{qi2019hybrid, wang2021modeling}. (c) Sparse distribution of the 152 ground monitoring stations. (d) The $0.25^{\circ}$ grid used for dense map inference, where most cells lack ground truth.}
    \label{fig:study_area}
\end{figure}

Our study targets the \textbf{BTHSA}, specifically anchoring on the strategic \textbf{"2+26" cities} air pollution transmission channel~\citep{lu2021estimation} (Figure~\ref{fig:study_area}). This region is characterized by highly heterogeneous topography, flanked by the \textbf{Taihang Mountains} to the west and the \textbf{Bohai Sea} to the east, enclosing the vast North China Plain. Such complex terrain induces distinct meteorological patterns (e.g., mountain blocking effects and land-sea circulation), resulting in \textbf{sharp pollution gradients} and frequent accumulation episodes. These challenging environmental dynamics render the BTHSA an ideal \textbf{testbed} for developing and validating the robustness of our physics-guided inference framework.

To comprehensively capture these spatiotemporal dynamics, we constructed a multi-source dataset integrating ground observations, meteorological reanalysis, and emission inventories. As summarized in Table~\ref{table:data_summary}, all datasets were processed to a unified hourly temporal resolution and spatially aligned to a uniform $0.25^{\circ} \times 0.25^{\circ}$ grid to support the grid inference task.

\begin{table*}[t!]
\centering
\caption{Summary of multi-source datasets used in BTHSA. Notations $\mathbf{X}$, $\mathbf{P}$, and $\mathbf{Q}$ denote Air Pollutants, Meteorology, and Emissions, respectively.}
\label{table:data_summary}
\resizebox{\textwidth}{!}{%
\begin{tabular}{@{} l l l c l @{}} 
\toprule
\textbf{Category (Notation)} & \textbf{Variable} & \textbf{Description} & \textbf{Unit} & \textbf{Source} \\
\midrule
\textbf{Target ($\mathbf{X}$)} & \PM & Fine particulate matter & $\mu\text{g/m}^3$ & CNEMC \\
\midrule
\multirow{5}{*}{\textbf{Meteorology ($\mathbf{P}$)}} 
 & $t2m, d2m$ & 2m Temperature \& Dewpoint & $K$ & \multirow{5}{*}{ERA5} \\
 & $rh2m, sp$ & Relative Humidity \& Surface Pressure & $\%, hPa$ & \\
 & $tp, blh$ & Total Precipitation \& Boundary Layer Height & $m$ & \\
 & $u_{100}, v_{100}$ & 100m U/V Wind Components & $m/s$ & \\
 & $rad$ & Surface Short-wave Radiation & $W/m^2$ & \\
\midrule
\multirow{2}{*}{\textbf{Emissions ($\mathbf{Q}$)}} 
 & Gases & \NitrogenOxides, \SulfurDioxide, \Ammonia, \VOC & $ton$ & \multirow{2}{*}{MEIC} \\
 & Particulates & primary \PM, \PMcoarse & $ton$ & \\
\midrule
\textbf{Auxiliary} & AOD & Aerosol Optical Depth & - & JMA (Himawari-8) \\
\bottomrule
\end{tabular}%
}
\end{table*}

\begin{itemize}
    \item \textbf{Air Pollutants ($\mathbf{X}$):} Hourly \PM\ concentrations were sourced from the China National Environmental Monitoring Center (CNEMC). Following rigorous quality control, data from 152 national monitoring stations (Figure~\ref{fig:study_area}(c)) were retained to serve as the ground truth.
    
    \item \textbf{Meteorology ($\mathbf{P}$):} Key meteorological variables governing pollutant transport and diffusion—including wind components ($u_{100}, v_{100}$), temperature ($t2m$), precipitation ($tp$), and boundary layer height ($blh$)—were sourced from the ERA5 reanalysis dataset. These variables represent the atmospheric conditions driving \PM\ formation and accumulation~\citep{chen2017detecting}.
    
    \item \textbf{Emissions ($\mathbf{Q}$):} Regional anthropogenic emissions of \NitrogenOxides, \VOC, \SulfurDioxide, \Ammonia, \PMcoarse, and \PM were obtained from the \textbf{Multi-resolution Emission Inventory for China (MEIC)}~\citep{li2017anthropogenic, geng2024efficacy}. While these monthly inventories are statistical estimates, they effectively capture general emission trends. To align with hourly air quality and meteorological data, we downscaled them to an hourly resolution using the temporal profile methodology described in the technical guideline~\citep{inventory}. Emission variables include primary \PM and \PMcoarse, which are distinct from the ambient concentration target.
    
    \item \textbf{Satellite AOD ($\mathbf{X}^{AOD}$):} Hourly AOD products from the Himawari-8 satellite were utilized as an auxiliary spatial constraint. While AOD is highly correlated with surface \PM, it suffers from frequent data missingness due to cloud cover, a challenge we specifically address via our proposed loss function.
\end{itemize}

\subsection{Problem Formulation: Inductive Spatiotemporal Kriging}
We formulate the inference task as an \textit{inductive spatiotemporal kriging} problem. The core objective is to learn a generalized mapping function capable of inferring \PM concentrations at arbitrary unobserved locations by leveraging sparse monitoring data and auxiliary physical fields. This problem is instantiated through two distinct downstream tasks:

\begin{itemize}
    \item \textbf{Station Inference:} Inferring time-series data for a held-out subset of monitoring stations based on the remaining observed stations. This task assesses the model's ability to interpolate and generalize within the sparse monitoring network.
    
    \item \textbf{Grid Inference:} Generating a continuous, high-resolution ($0.25^{\circ}$) \PM spatial field by treating all grid cells without stations as unobserved target nodes. This task aims to reconstruct the full spatial distribution of pollutants to eliminate monitoring blind spots.
\end{itemize}

Formally, let $G = (\mathcal{V}, \mathcal{E})$ denote a graph where $\mathcal{V}$ represents the set of all spatial locations (stations or grid cells). The node set is dynamically partitioned into observed nodes $\mathcal{V}_{obs}$ and unobserved target nodes $\mathcal{V}_{target}$. Given the historical pollutant observations at $\mathcal{V}_{obs}$, along with meteorological conditions $\mathbf{P}$ and emissions $\mathbf{Q}$ for all nodes, we aim to learn a mapping function $\mathcal{F}$ to infer \PM concentrations at $\mathcal{V}_{target}$:
\begin{equation}
    \mathbf{\hat{X}}_{target} = \mathcal{F}(\mathbf{X}_{obs}, \mathbf{P}, \mathbf{Q}, G)
\end{equation}

A critical challenge in \textbf{Grid Inference} is ensuring the physical plausibility of the generated field over vast unmonitored regions. While satellite-derived Aerosol Optical Depth (AOD) offers valuable spatial guidance, it suffers from high spatiotemporal sparsity due to cloud cover and nighttime limitations. To address this, our framework introduces AOD data ($\mathbf{X}^{AOD}$) not as a direct input feature, but as a \textit{spatial structural constraint} during training. This strategy allows the model to leverage satellite spatial patterns while effectively overcoming the pervasive data missingness challenge.

\subsection{The \ourmodel\ Architecture}
\begin{figure}[H]
    \centering
    \includegraphics[width=1.0\textwidth]{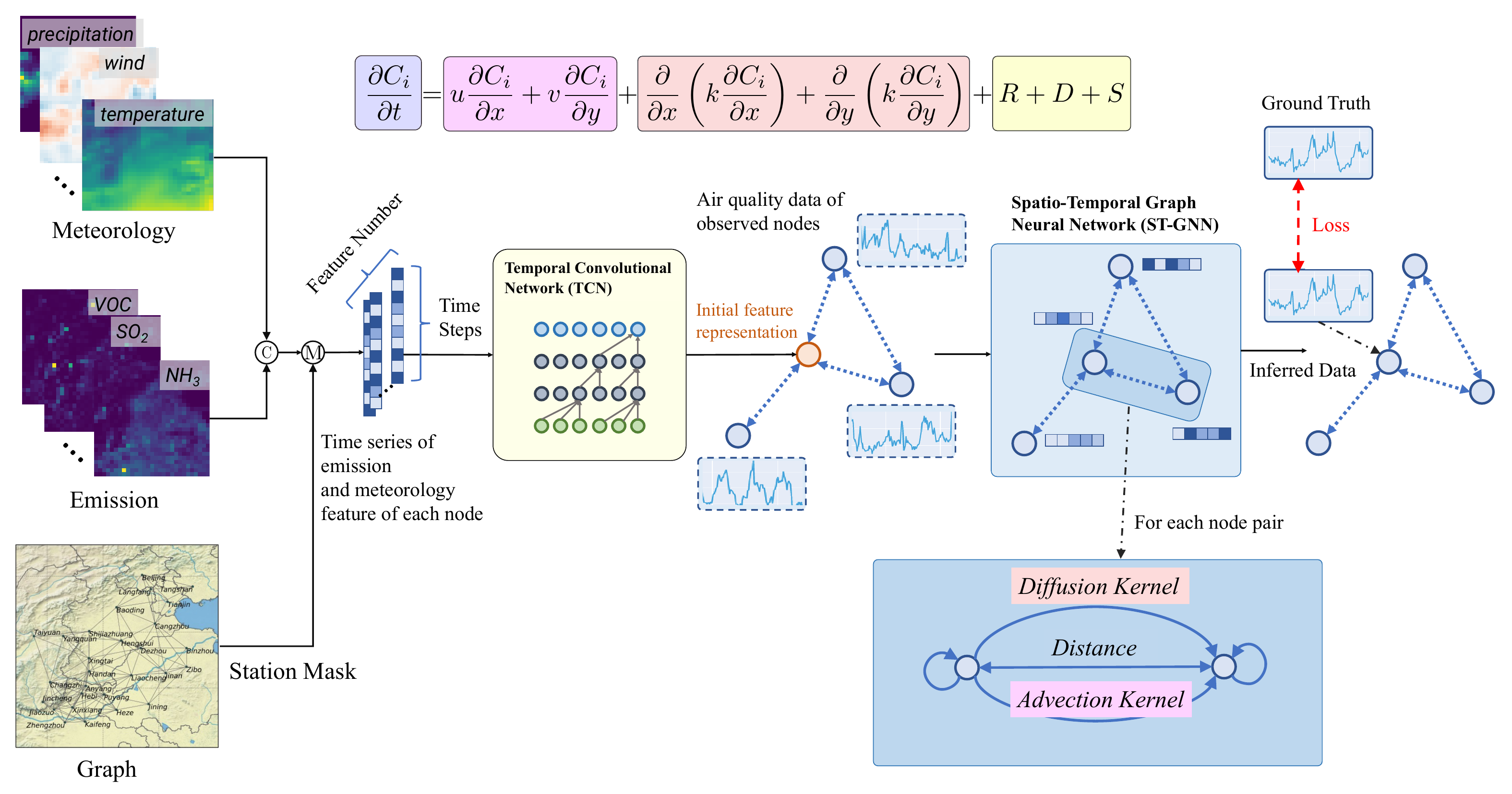}
    \caption{\textbf{Architecture of \ourmodel.} The model first encodes local temporal dynamics via TCN. It then propagates spatial information via a Physics-Guided Graph Network, driven by two explicit kernels: a \textit{Diffusion Kernel} (isotropic spreading) and an \textit{Advection Kernel} (wind-driven transport). Finally, the training is constrained by a novel Masked AOD Gradient Loss to ensure structural consistency with satellite observations.}
    \label{fig:model_architecture}
\end{figure}

The proposed \ourmodel\ (Figure~\ref{fig:model_architecture}) embodies a "Physics-Embedded" design philosophy, seamlessly integrating deep representation learning with explicit atmospheric dynamic laws. The architecture consists of three hierarchical stages: temporal feature encoding, physics-guided propagation, and constrained optimization.

\subsubsection{Temporal Feature Encoding (TCN)}
In the first stage, we employ a Temporal Convolutional Network (TCN) to encode the local temporal dynamics for each node. Let $\mathcal{X}_i^{in} = [\mathbf{P}_i, \mathbf{Q}_i]$ denote the input time-series of meteorology and emissions for node $i$. The TCN maps this sequence to a high-dimensional latent representation $\mathbf{H}_i^{(0)}$:
\begin{equation}
    \mathbf{H}_i^{(0)} = \text{TCN}(\mathcal{X}_i^{in})
\end{equation}
Compared to RNNs, TCNs employ dilated causal convolutions to capture long-range temporal dependencies (e.g., cumulative emission effects) more efficiently with parallel computing~\citep{bai2018empirical}. This step provides a physics-informed initialization for all nodes, offering a robust starting point even for unmonitored grid cells where no historical pollution data is available.

\subsubsection{Physics-Guided Spatiotemporal Propagation}
A single graph structure is insufficient to capture the multifaceted interactions in atmospheric systems. Therefore, we explicitly model the spatiotemporal dependencies using a set of three distinct graphs, each encoding a specific physical or statistical relationship:

\textbf{1. Geospatial Graph ($G^{\mathcal{S}}$):} This graph captures the fundamental principle of spatial autocorrelation (Tobler's First Law of Geography). The adjacency matrix $A^{\mathcal{S}}$ is constructed based on the pairwise geographic distances between nodes $v_i$ and $v_j$ using a Gaussian kernel, thresholded by a distance $\xi$:
\begin{equation}
    A_{ij}^{\mathcal{S}} = \begin{cases} 
        \exp\left(-\frac{\text{dist}(v_i, v_j)^2}{\sigma^2}\right), & \text{if dist}(v_i, v_j) < \xi \\ 
        0, & \text{otherwise} 
    \end{cases}
    \label{eq:adjacency_S}
\end{equation}
where $\sigma^2$ is the variance of distances. This graph serves as the structural basis for modeling isotropic interactions.

\textbf{2. Diffusion Graph ($G^{\mathcal{D}}$):} Based on $G^{\mathcal{S}}$, this graph models the physical process of diffusion, where pollutants spread from high to low concentration areas. Mathematically, this corresponds to the Laplacian operator on the graph. We define the diffusion operator $\tilde{A}^{\mathcal{D}}$ by symmetrically normalizing $A^{\mathcal{S}}$:
\begin{equation}
    \tilde{A}^{\mathcal{D}} = \mathbf{D}_{\mathcal{S}}^{-\frac{1}{2}} A^{\mathcal{S}} \mathbf{D}_{\mathcal{S}}^{-\frac{1}{2}}
    \label{eq:adjacency_D}
\end{equation}
where $\mathbf{D}_{\mathcal{S}}$ is the degree matrix of $A^{\mathcal{S}}$. This operator allows the GNN to simulate the smoothing effect of atmospheric diffusion.

\begin{figure}[H]
    \centering
    \includegraphics[width=0.9\textwidth]{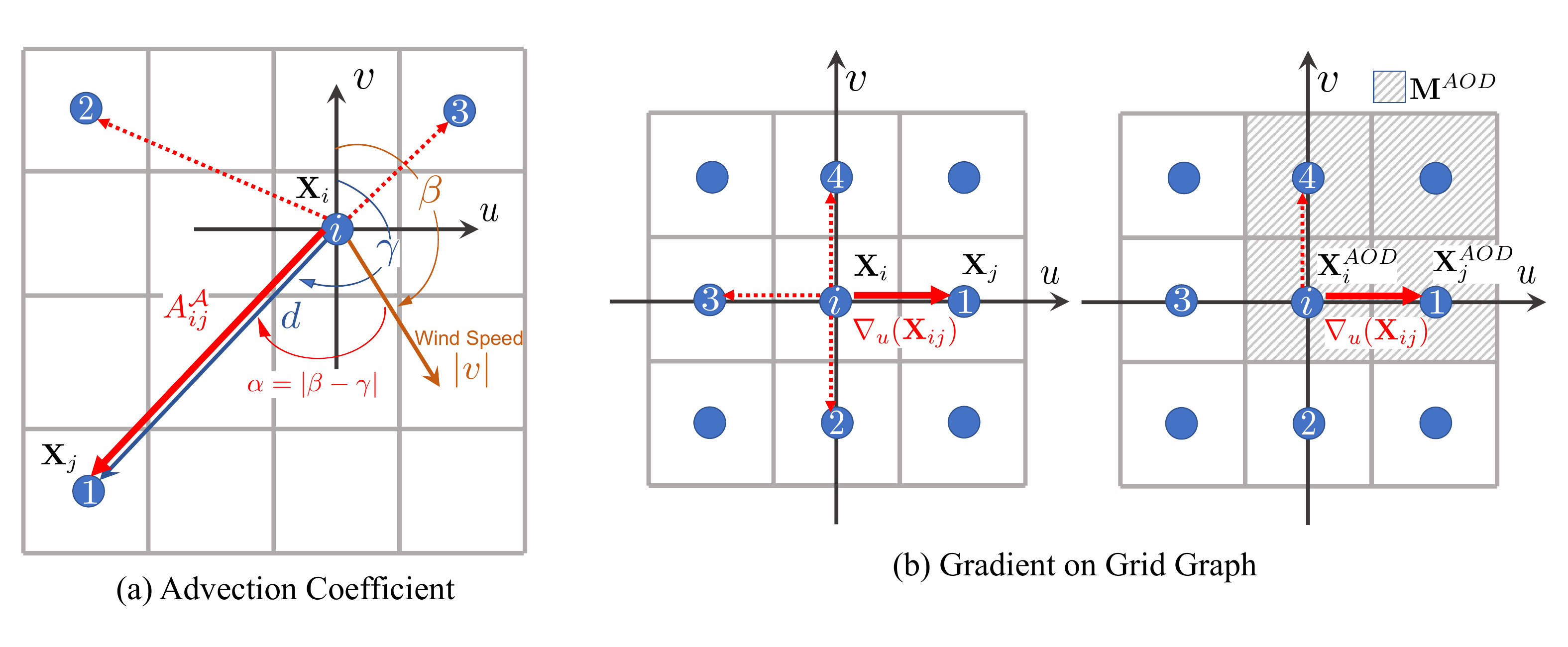}
    \caption{\textbf{Physical Kernels and AOD Constraint.} (a) The \textbf{Advection Kernel} weights edges based on wind velocity projections, modeling directional transport. (b) The \textbf{AOD Gradient Loss} aligns the spatial difference of predictions with AOD, active only on valid pixels (shaded area), ensuring robustness to missing data.}
    \label{fig:kernels}
\end{figure}

\textbf{3. Advection Graph ($G^{\mathcal{A}}$):} This kernel explicitly models anisotropic transport driven by wind fields. A directed edge exists from node $j$ to $i$ only if $j$ is upwind. As visually illustrated in Figure~\ref{fig:kernels}(a), the edge weight is determined by the projection of the wind vector $\vec{v}$ onto the node-pair vector $\vec{e}_{ji}$:
\begin{equation}
    A_{ij}^{\mathcal{A}} = \begin{cases} 
        \text{ReLU}\left( \frac{|\vec{v}|}{d_{ij}} \cdot \cos(\alpha) \right), & \text{if } d_{ij} < \xi \\
        0, & \text{otherwise}
    \end{cases}
    \label{eq:adjacency_A}
\end{equation}
where $\alpha$ is the angle between the wind direction and the vector connecting node $j$ to $i$, and $d_{ij}$ is the distance. This kernel encodes the physical prior that pollutants transfer significantly faster and stronger along the wind direction.

\textbf{Propagation Layer:} The node representation $\mathbf{H}^{(l)}$ at layer $l$ is updated by aggregating messages from both the Diffusion and Advection operators:
\begin{equation}
    \mathbf{H}^{(l)} = \sigma \left( \left( \tilde{A}^{\mathcal{D}} \mathbf{H}^{(l-1)} + A^{\mathcal{A}} \mathbf{H}^{(l-1)} \right) \mathbf{W}^{(l)} \right)
    \label{eq:gnn_layer}
\end{equation}
where $\mathbf{W}^{(l)}$ is the learnable weight matrix and $\sigma$ is the activation function. This parallel propagation disentangles the isotropic diffusion (local spreading) and anisotropic advection (directional transport), enabling the model to learn physically interpretable patterns.

\subsubsection{Readout Layer and Final Prediction}
The final node representations $\mathbf{H}^{(L)}$ synergize \textit{local temporal dynamics} (encoded by TCN) with \textit{spatial context} aggregated from neighbors via physics-guided propagation. Since target nodes lack historical ground truth, we employ a node-wise shared Multi-Layer Perceptron (MLP) to directly map these integrated features to the pollutant concentration:
\begin{equation}
    \mathbf{\hat{X}}_i^t = \text{MLP}(\mathbf{H}_i^{(L), t})
\end{equation}
This mapping enables inference based on the full spatiotemporal context—local interactions combined with regional transport—effectively filling data gaps in unobserved regions.

\subsection{Training Strategy}

\subsubsection{Inductive Node Masking}
To enforce inductive generalization, we employ a dynamic masking strategy~\citep{wu2021inductive}. In each training iteration, we randomly mask a subset (e.g., 50\%) of the observed stations as "unknown" target nodes. The model is forced to reconstruct the values of these masked nodes by combining their \textbf{local meteorological and emission features} with \textbf{pollution context propagated from the remaining observed nodes}. This mechanism prevents the model from overfitting to specific station IDs and accurately simulates the real-world scenario of inferring unmonitored locations where historical air quality data is absent.

\subsubsection{Composite Loss with AOD Gradient Constraint}
A critical innovation of \ourmodel\ is the integration of multi-stage supervision and satellite constraints. We define a composite loss function $\mathcal{L}$:
\begin{equation}
    \mathcal{L} = \mathcal{L}_{infer} + \lambda_1 \mathcal{L}_{init} + \lambda_2 \mathcal{L}_{AOD}
\end{equation}
where $\lambda_1$ and $\lambda_2$ are hyperparameters balancing the contribution of initialization and spatial constraints.

\textbf{1. Inference Loss ($\mathcal{L}_{infer}$):}
This is the primary objective. After the physics-guided GNN aggregates real pollution information from observed neighbors to refine the initial features, the final prediction $\hat{X}$ is generated. We minimize the L1 error on the \textbf{masked target nodes} ($\mathcal{V}_{target}$) to simulate the inference scenario:
\begin{equation}
    \mathcal{L}_{infer} = \sum_{t=1}^{T} \sum_{i \in \mathcal{V}_{target}} \left| \hat{X}_i^t - X_i^t \right|
\end{equation}
This ensures the model learns to accurately infer unknown locations by combining propagated neighbor information with local features.

\textbf{2. Initialization Loss ($\mathcal{L}_{init}$):}
To enable robust inference, we require the TCN to generate a reasonable "initial proposal" ($\hat{X}^{init}$) based solely on the node's local meteorology and emissions, \textit{before} any neighbor information is aggregated. Crucially, this loss is also calculated on the \textbf{target nodes}:
\begin{equation}
    \mathcal{L}_{init} = \sum_{t=1}^{T} \sum_{i \in \mathcal{V}_{target}} \left| \text{MLP}(\mathbf{H}_i^{(0), t}) - X_i^t \right|
\end{equation}
This constraint forces the TCN to act as a solid "prior" generator. Even without neighbor information, the model can produce a physics-based baseline estimate, which is then refined by the GNN layers. This follows the inductive training paradigm described in \citet{wu2021inductive}.

\textbf{3. Masked AOD Spatial Gradient Loss ($\mathcal{L}_{AOD}$):}
This term introduces spatial structure constraints, particularly for Grid Inference. We align the \textit{spatial gradient} of the final prediction with that of the satellite AOD field. A binary mask $\mathbf{M}$ ensures robustness against missing satellite data:
\begin{equation}
    \mathcal{L}_{AOD} = \sum_{t=1}^{T} \sum_{(i,j) \in \mathcal{E}} \left| \nabla_{ij}(\mathbf{\hat{X}}) - \nabla_{ij}(\mathbf{X}^{AOD}) \right| \cdot \mathbf{M}_{ij}^t
\end{equation}
As visualized in Figure~\ref{fig:kernels}(b), $\nabla_{ij}(\mathbf{X}) = X_j - X_i$ approximates the spatial gradient between adjacent nodes. This loss injects satellite-derived structural knowledge (e.g., pollution plume shapes) into the model without requiring AOD as a direct input, ensuring continuous and physically consistent mapping.

\section{Results}
\label{sec:results}

To comprehensively validate the \textbf{inductive generalization} and \textbf{physical consistency} of \ourmodel, the experimental evaluation is structured into three hierarchical stages. First, we establish the model's quantitative superiority in \textit{Station Inference} against state-of-the-art baselines. Second, we conduct a "stress test" under extreme data scarcity (50\% missing stations) to evaluate robustness. Finally, we provide a forensic mechanism analysis of \textit{Grid Inference}, demonstrating how the physics-informed kernels and AOD constraints cooperate to generate continuous, high-fidelity pollution maps.

\subsection{Quantitative Superiority in Station Inference}
We first evaluate the core inductive capability by partitioning the 152 monitoring stations into a training set (70\%, 106 stations) and a \textbf{completely unobserved} test set (30\%, 46 stations). To rigorously prevent temporal data leakage, we employed a strict chronological split strategy:
\begin{itemize}
    \item \textbf{General Evaluation:} To strictly prevent temporal data leakage, data from \textbf{2016 to 2018} were used for training, \textbf{2019} for validation, and \textbf{2020} for testing.
    \item \textbf{Seasonal Evaluation:} To assess adaptability to extreme seasonal variations, we constructed specific sub-datasets:
    \begin{itemize}
        \item \textbf{Winter (Heating Season):} Defined as November to March. The training set includes winters from Jan 2016 to Mar 2018, while the validation and test sets cover the winters of \textbf{2018/2019} (Nov 2018–Mar 2019) and \textbf{2019/2020} (Nov 2019–Mar 2020), respectively.
        \item \textbf{Summer (Warm Season):} Defined as May to September. The training set spans the summers of \textbf{2016 to 2018}, with validation and testing performed on the summers of \textbf{2019} and \textbf{2020}, respectively.
    \end{itemize}
\end{itemize}

Table~\ref{tab:performance_comparison} summarizes the performance benchmarks. While spatiotemporal models generally outperform traditional feature-based and purely temporal methods, \ourmodel\ establishes a new state-of-the-art.
\begin{itemize}
    \item \textbf{Overall Performance:} \ourmodel\ achieves a yearly Mean Absolute Error (MAE) of \textbf{9.52 \ug}, representing a substantial \textbf{25.2\% improvement} over the strongest baseline, IGNNK~\citep{wu2021inductive}. This confirms that explicitly modeling physical transport yields superior generalization compared to purely statistical interpolation (Kriging).
    
    \item \textbf{Seasonal Robustness (The Winter Challenge):} The performance gap is most pronounced in winter (MAE 15.09 \ug), a season characterized by severe pollution episodes and complex meteorological conditions. Traditional models like STGCN struggle here (MAE 23.43 \ug). The fundamental limitation is that STGCN relies on \textbf{static graphs based solely on geographic distance}, implicitly assuming isotropic influence between locations. It fails to capture the \textbf{anisotropic} nature of wind-driven transport (i.e., upwind sources affect downwind targets, not vice-versa), which is the dominant driver of winter pollution. In contrast, \ourmodel\ utilizes the dynamic \textit{Advection Kernel} to explicitly model this directional transport, ensuring robust inference.
    
    \item \textbf{Baseline Analysis:} Feature-based models (e.g., XGBoost) and temporal models (e.g., LSTM) fail to capture spatial correlations, resulting in high errors ($>20$ \ug). Even advanced inductive baselines like IGNNK, while capable of dynamic graph modeling, still construct connections based on \textbf{spatial proximity}. Both STGCN and IGNNK lack an explicit mechanism to encode \textbf{physical dynamics} (e.g., diffusion vs. advection). \ourmodel's success validates that embedding domain knowledge—specifically separating wind direction from distance—provides a stronger inductive bias than learning from proximity-based data alone.
\end{itemize}

\begin{table}[H]
\centering
\caption{\textbf{Comparison of \PM inference performance on unobserved stations (30\% hold-out).} \ourmodel\ achieves the lowest error (MAE) across all seasons, demonstrating superior inductive generalization compared to feature-based, temporal, and data-driven spatiotemporal baselines.}
\label{tab:performance_comparison}
\resizebox{0.8\textwidth}{!}{%
\begin{tabular}{@{}ll|c|c|c@{}}
\toprule
\multirow{2}{*}{\textbf{Model Category}} & \multirow{2}{*}{\textbf{Methods}} & \textbf{All Year} & \textbf{Winter} & \textbf{Summer} \\
 &  & MAE$\downarrow$ & MAE$\downarrow$ & MAE$\downarrow$ \\ \midrule
\multirow{2}{*}{Feature-based} & XGBoost & 24.21 & 35.97 & 15.49 \\
 & MLP & 25.71 & 38.67 & 15.30 \\ \midrule
\multirow{2}{*}{Temporal} & LSTM & 24.65 & 34.96 & 14.49 \\
 & GRU & 22.79 & 34.17 & 15.00 \\ \midrule
\multirow{2}{*}{Spatiotemporal} & STGCN & 17.84 & 23.43 & 11.23 \\
 & IGNNK & 12.73 & 16.29 & 8.53 \\ \midrule
\textbf{Physics-informed} & \textbf{\ourmodel} (Ours) & \textbf{9.52} & \textbf{15.09} & \textbf{7.65} \\ \bottomrule
\end{tabular}%
}
\end{table}

\subsection{Robustness Verification under Sparse Observation}
To probe the limits of \ourmodel\ in data-scarce scenarios, we increased the difficulty by randomly masking \textbf{50\%} of the stations. This "stress test" evaluates whether the physics-informed kernels can maintain structural integrity when local information is severely depleted. The evaluation is conducted across three dimensions: regional consistency, error bounds analysis, and edge-case stability.

\begin{figure}[H]
    \centering
    \includegraphics[width=\textwidth]{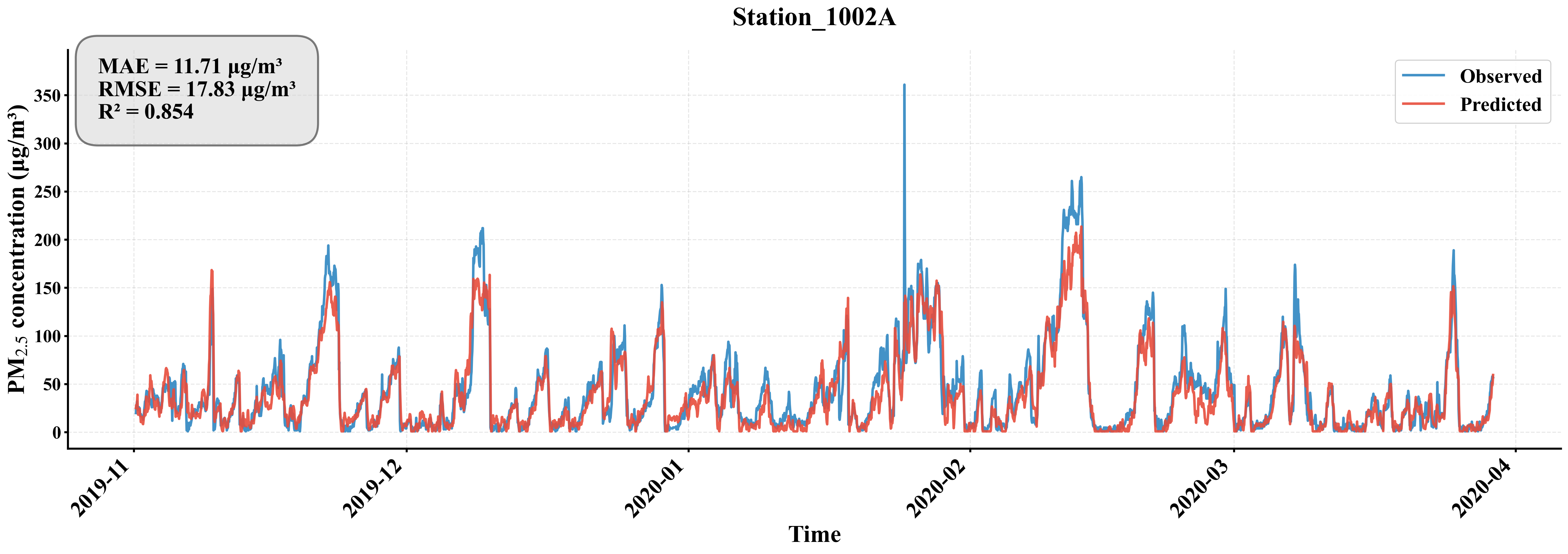}\\[0.35em]
    \includegraphics[width=\textwidth]{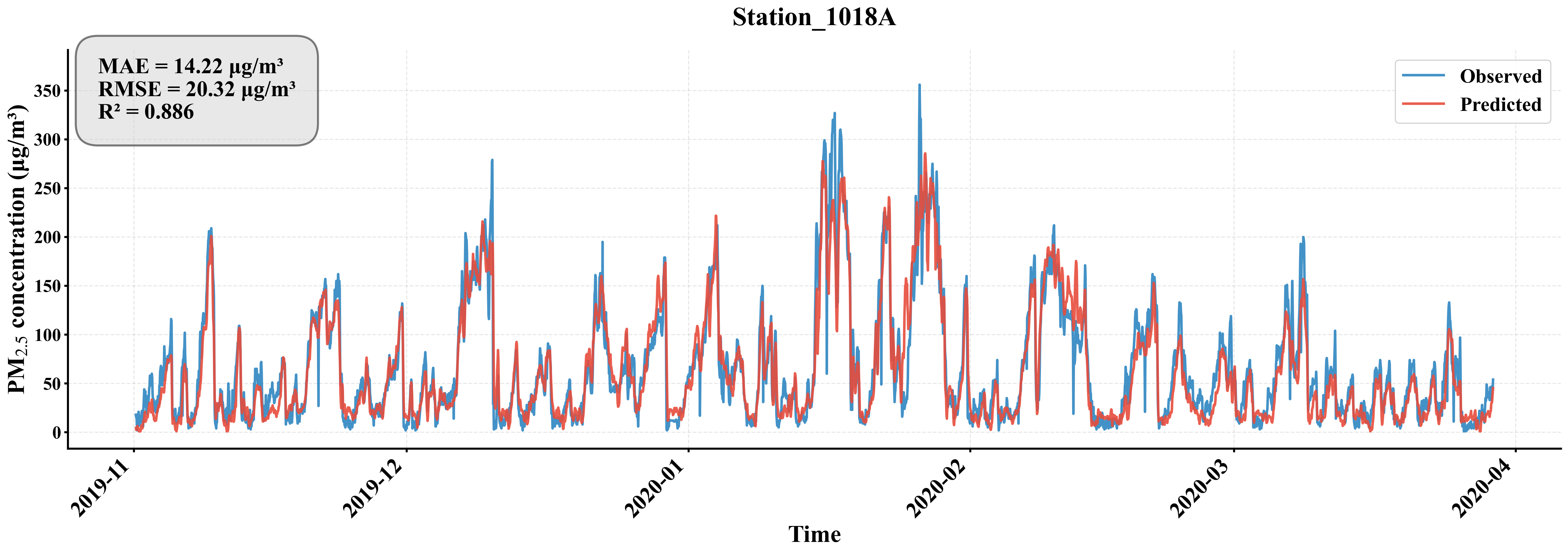}\\[0.35em]
    \includegraphics[width=\textwidth]{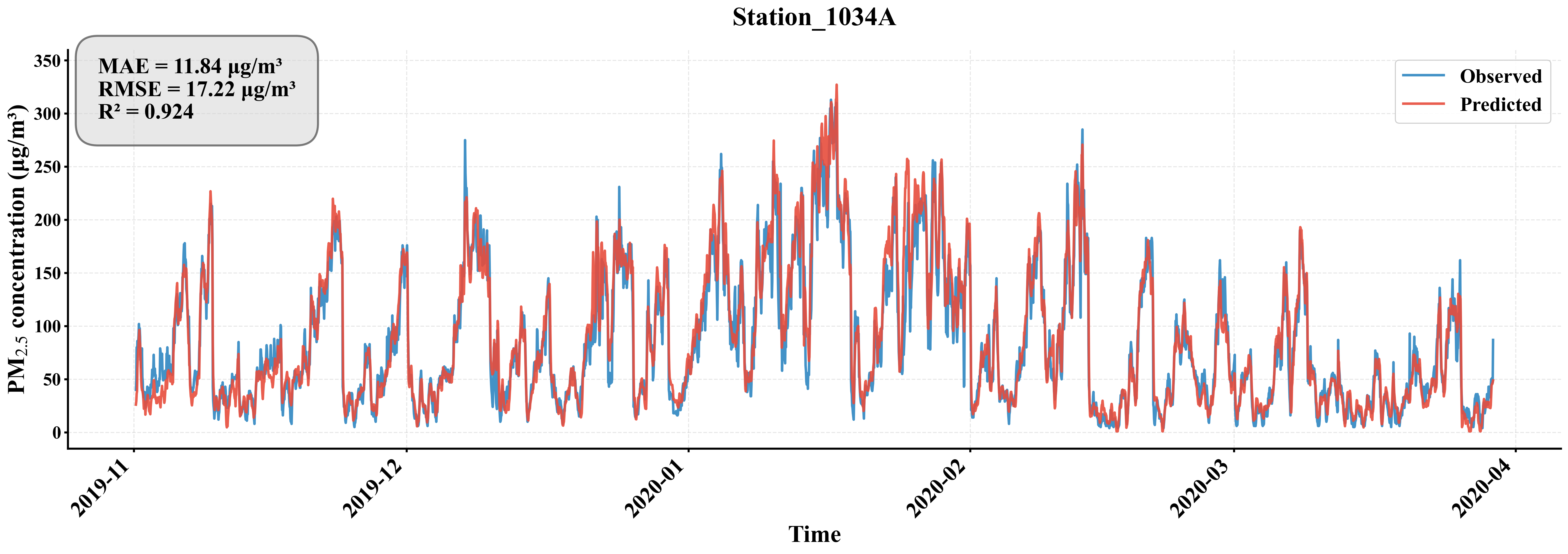}
    \caption{\textbf{Inference performance across core cities in the BTHSA region.} From top to bottom: \textbf{Beijing} (Station 1002A), \textbf{Tianjin} (Station 1018A), and \textbf{Shijiazhuang} (Station 1034A). Despite being completely unobserved during training, the model accurately tracks the ground truth (blue) with high correlation ($R^2 > 0.85$) across these geographically distinct urban centers, successfully capturing the complex winter pollution cycles.}
    \label{fig:bj_group}
\end{figure}

\textbf{1. Regional Generalization (Central Urban Agglomeration):}
Figure~\ref{fig:bj_group} illustrates inference results for three representative unobserved stations in \textbf{Beijing, Tianjin, and Shijiazhuang}—the core cities of the BTHSA pollution transmission channel. Even with half the network missing, the model accurately reconstructs synoptic-scale variability ($R^2 > 0.85$) across these geographically distinct centers. It captures both the rapid accumulation and dissipation phases of regional episodes (e.g., late January 2020), demonstrating that the learned physical kernels effectively generalize across the major transport corridor.

\begin{figure}[H]
    \centering
    \includegraphics[width=\textwidth]{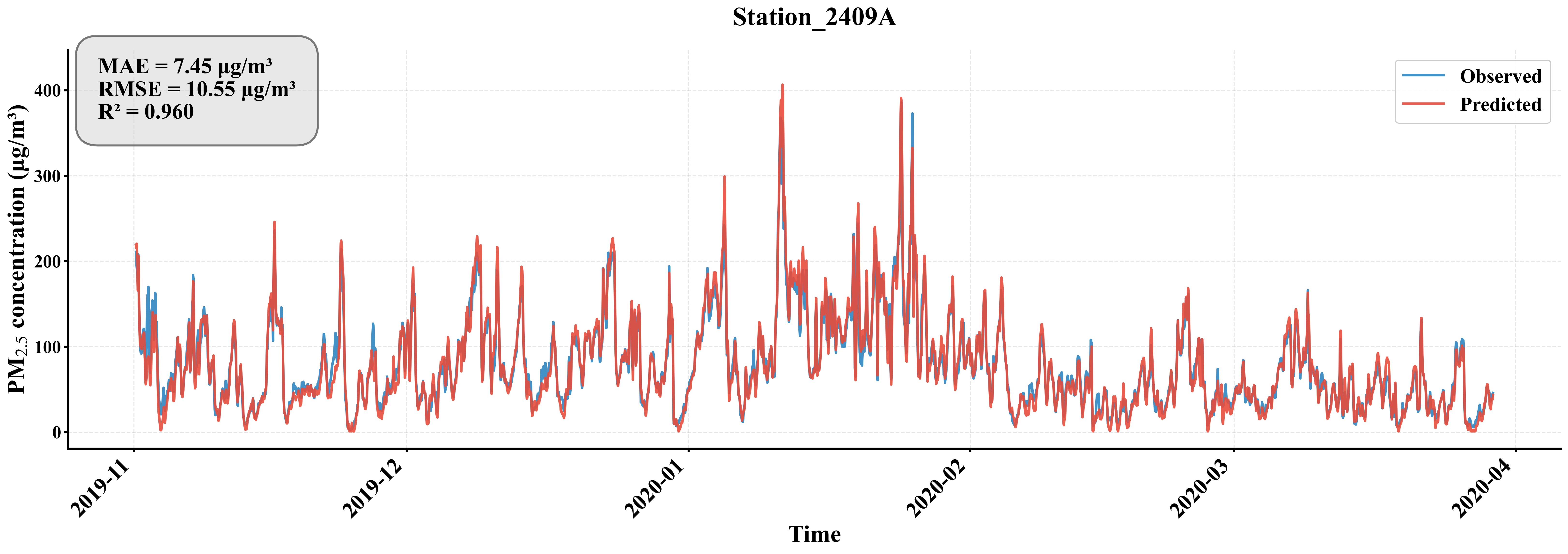}\\[0.35em]
    \includegraphics[width=\textwidth]{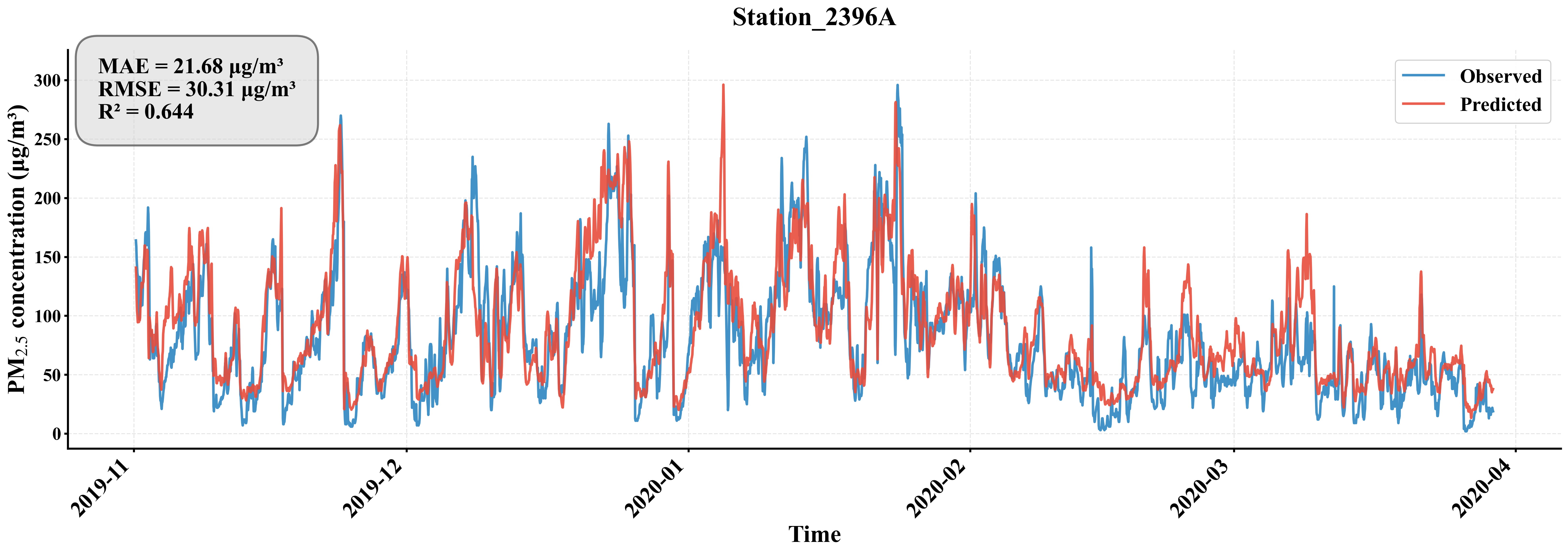}
    \caption{\textbf{Performance bounds analysis.} Top: The best-performing case (Station 2409A) shows near-perfect reconstruction due to dense neighbor support. Bottom: The worst-case scenario (Station 2396A) still maintains correct trend directionality, demonstrating stability under challenging topology.}
    \label{fig:rmse_extremes}
\end{figure}

\textbf{2. Error Analysis (Topology Dependence):}
Figure~\ref{fig:rmse_extremes} provides a forensic contrast between best and worst-performing cases, revealing the influence of local station topology:
\begin{itemize}
    \item \textbf{Best Case (Station 2409A):} Located in a dense cluster, this station achieves near-perfect alignment ($R^2=0.96$). The dense connectivity allows the \textit{Diffusion Kernel} to effectively aggregate gradients, resulting in high-fidelity reconstruction.
    \item \textbf{Worst Case (Station 2396A):} This station exhibits larger residuals ($R^2=0.644$) during rapid shifts, likely due to micro-scale emission bursts smoothed out by neighbor averaging. However, the model still correctly identifies the \textit{onset timing} and \textit{trend direction}, proving that the \textit{Advection Kernel} successfully captures macro-scale transport even when local magnitude is underestimated.
\end{itemize}

\begin{figure}[H]
    \centering
    \includegraphics[width=\textwidth]{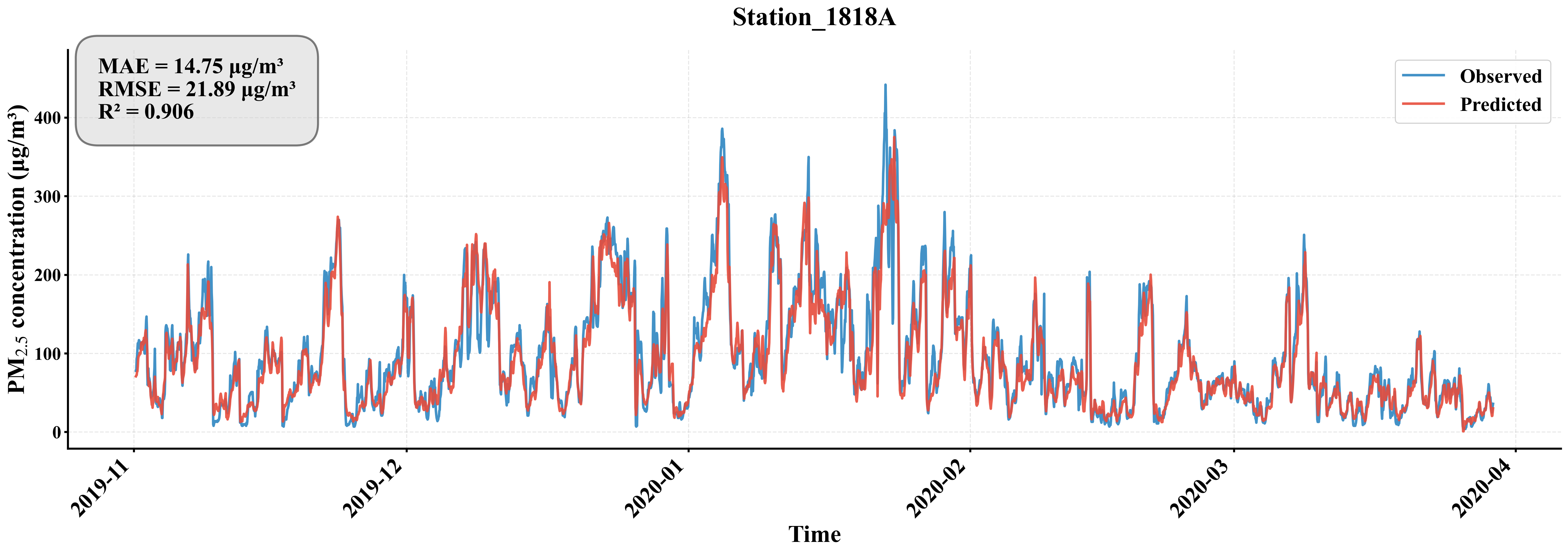}\\[0.35em]
    \includegraphics[width=\textwidth]{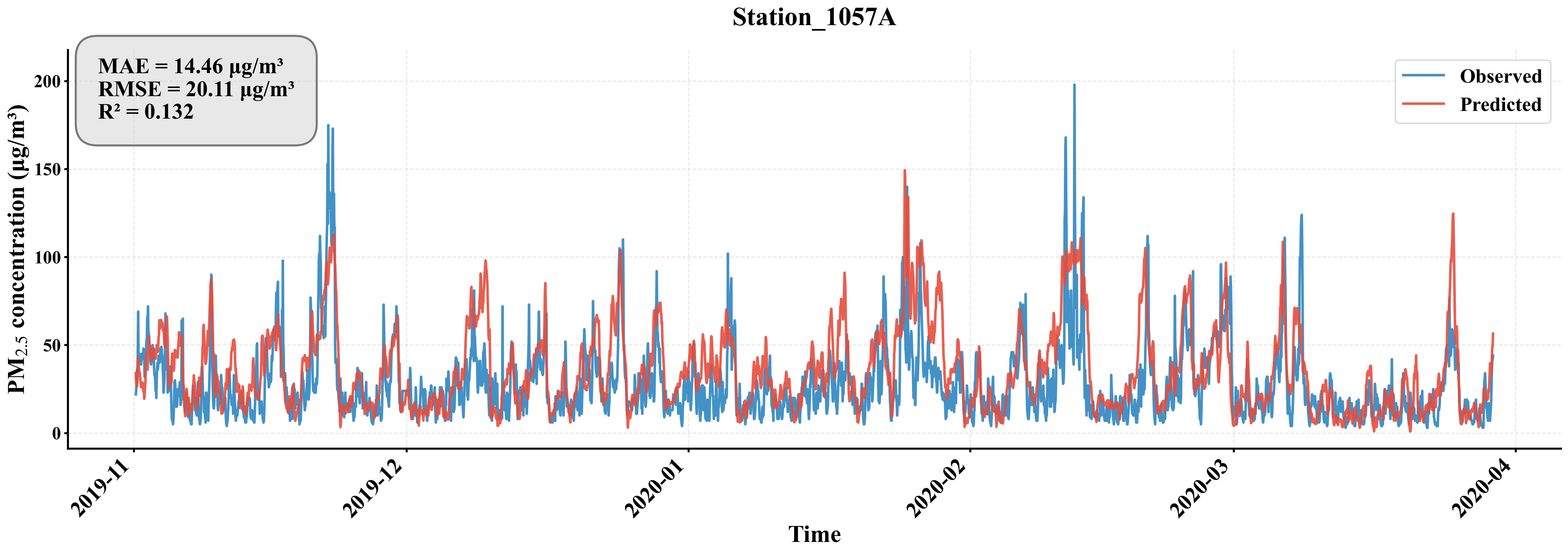}
    \caption{\textbf{Inference in spatially isolated regions.} Station 1057A (bottom) represents an extreme challenge with minimal upstream sensors. While high-frequency details are lost (lower $R^2$), the physics-based kernels ensure the inferred baseline trend remains physically plausible, acting as a conservative filter against hallucination.}
    \label{fig:sparse_neighbors}
\end{figure}

\textbf{3. Extreme Sparsity (Physics as a Safety Mechanism):}
A critical test is shown in Figure~\ref{fig:sparse_neighbors}, focusing on the challenging Station 1057A (Zhangjiakou), an edge node with almost no upstream sensors. While the quantitative metric drops ($R^2=0.132$) due to the loss of high-frequency variance, a significant physical property emerges: the model successfully recovers the \textit{low-frequency trend} and correct background concentration levels (20-50 \ug). It does not "hallucinate" zero values or random noise. This validates that our \textit{Diffusion Kernel} acts as a conservative low-pass filter in data vacuums, ensuring inference remains physically plausible even when information is extremely scarce.

\subsection{AOD-Guided Grid Inference: Mechanism Analysis}
\begin{figure}[H]
    \centering
    \begin{subfigure}{\textwidth}
        \centering
        \includegraphics[width=0.85\textwidth]{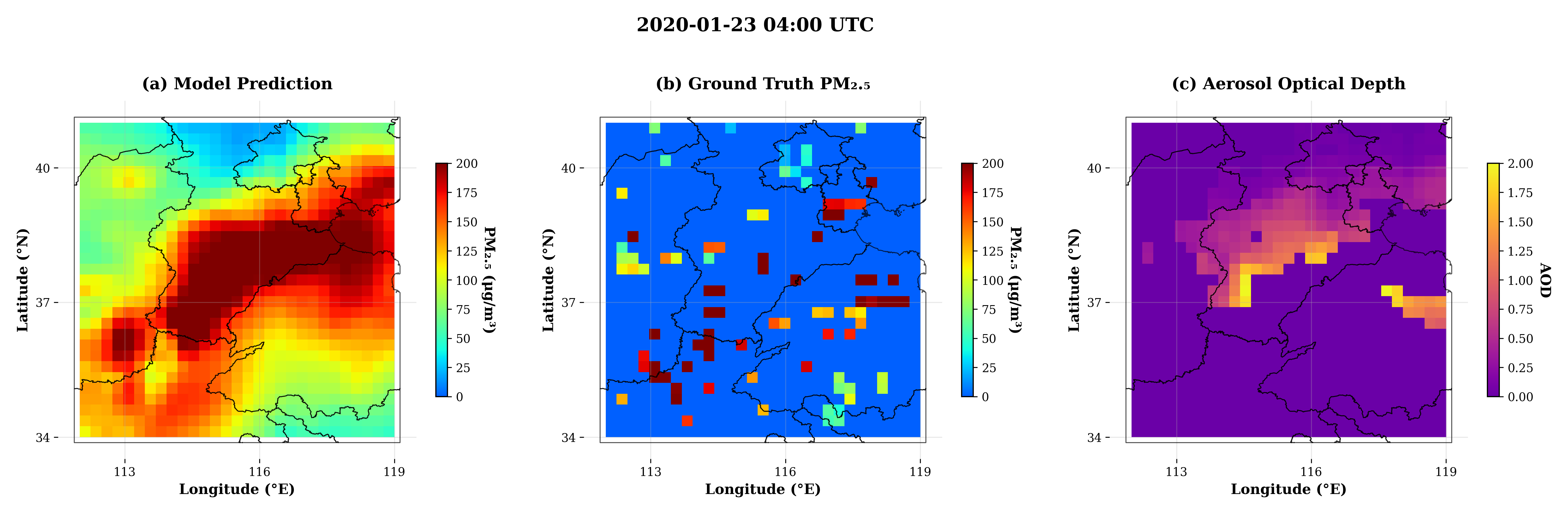}
    \end{subfigure}
    \begin{subfigure}{\textwidth}
        \centering
        \includegraphics[width=0.85\textwidth]{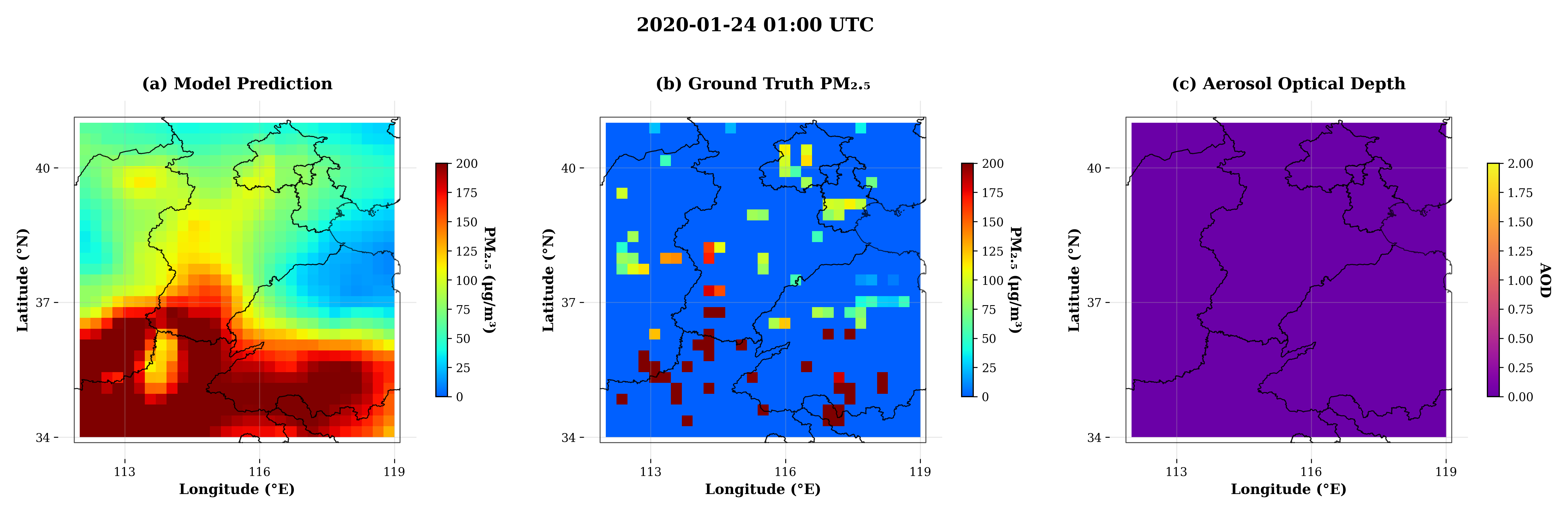}
    \end{subfigure}
    \begin{subfigure}{\textwidth}
        \centering
        \includegraphics[width=0.85\textwidth]{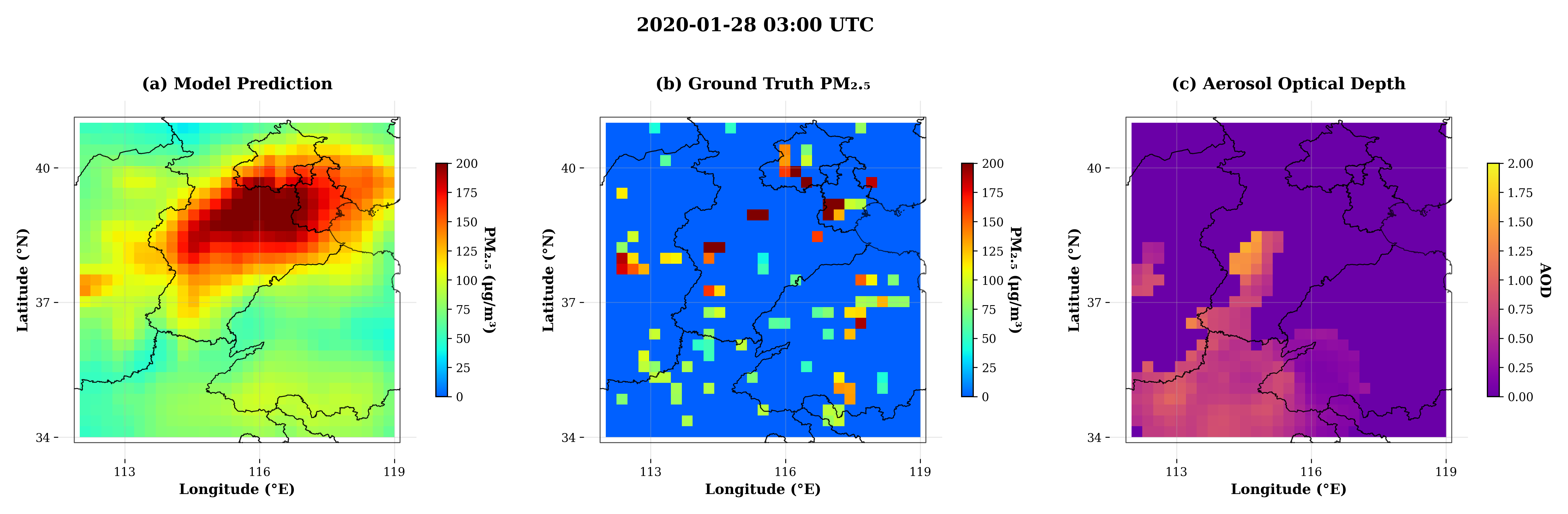}
    \end{subfigure}
    \begin{subfigure}{\textwidth}
        \centering
        \includegraphics[width=0.85\textwidth]{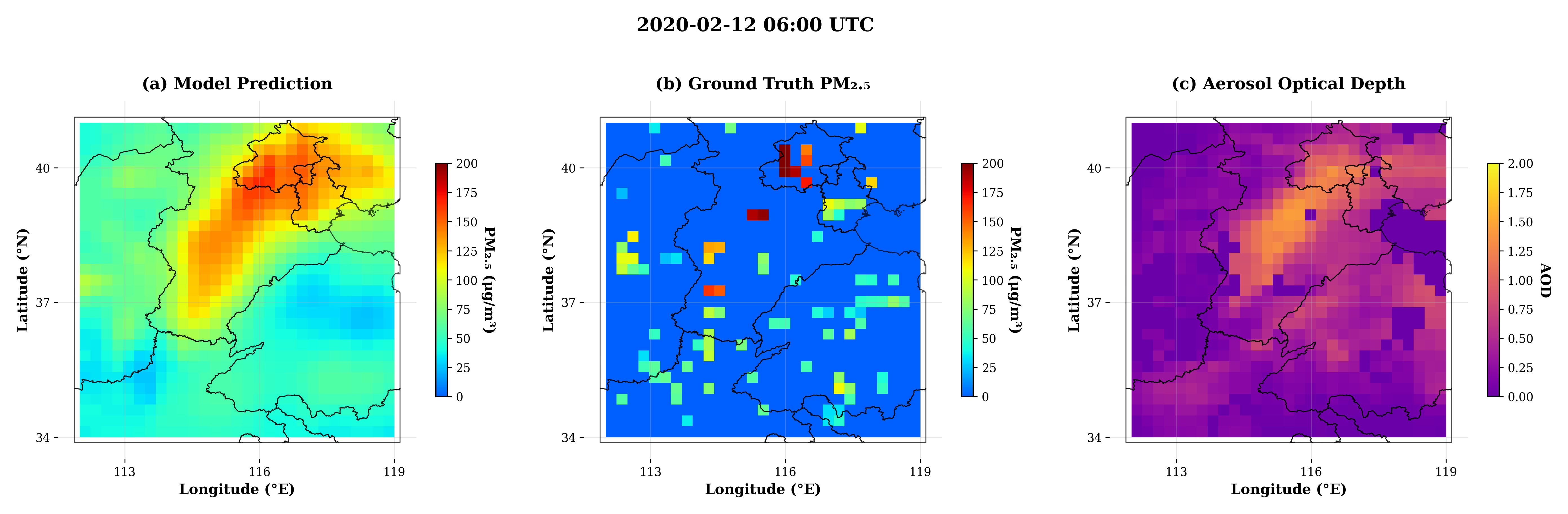}
    \end{subfigure}
    \caption{\textbf{Robustness analysis of Grid Inference across varying satellite data conditions.} Columns from left to right: Model Prediction, Ground Truth Stations, and Reference Satellite AOD. Note that AOD is used only for training constraints, not as a direct input.
    \textbf{(a) Ideal Case:} When AOD and ground truth are consistent, the model accurately reconstructs the plume structure.
    \textbf{(b) Missing AOD:} In the complete absence of satellite observations (e.g., nighttime or heavy cloud cover), the model robustly generates a smooth field driven by physical diffusion and advection kernels.
    \textbf{(c) Conflicting Data:} The model resists artifacts from misleading AOD patterns by anchoring to ground truth.
    \textbf{(d) Biased AOD:} The model adopts the correct spatial shape from AOD but corrects its magnitude bias.}
    \label{fig:grid_inference}
\end{figure}

The ultimate objective of \ourmodel\ is to generate continuous, high-resolution \PM\ maps (Grid Inference) that are both physically plausible and consistent with sparse ground observations. Figure~\ref{fig:grid_inference} presents a forensic analysis of the model's inference capabilities across four distinct scenarios. 

It is crucial to clarify that in our framework, \textbf{AOD is not used as a direct input feature} during inference; rather, it serves as a spatial gradient constraint during training. Therefore, the AOD panels in Figure~\ref{fig:grid_inference} act as an external reference to validate whether the model has successfully internalized the spatial structural patterns of pollution.

\begin{itemize}
    \item \textbf{Scenario (a): Ideal Consistency (Structure Transfer).} 
    In this baseline scenario, the satellite AOD field shows a clear pollution plume that is structurally consistent with the sparse ground measurements. \ourmodel\ successfully reproduces this complex spatial pattern, generating smooth transitions between stations. This indicates that the \textit{Masked AOD Spatial Gradient Loss} has effectively taught the model to bridge sparse points using realistic, satellite-derived spatial gradients when data sources align.

    \item \textbf{Scenario (b): Total Data Absence (Physics Fallback).} 
    This case represents a "nighttime" or "heavy cloud" scenario where AOD data is completely unavailable (shown as a blank field). A key advantage of our framework is that it does not "crash" or output artifacts in the absence of satellite data. Instead, the model robustly falls back to its internal physics-informed kernels—specifically the \textit{Advection} and \textit{Diffusion} operators. Driven by meteorological fields and ground stations alone, \ourmodel\ generates a physically smooth and continuous pollution field, demonstrating its capability to function independently of satellite inputs.

    \item \textbf{Scenario (c): Structural Conflict (Robustness).} 
    Here, the AOD data exhibits a spatial trend that conflicts with the high-fidelity ground observations (likely due to surface reflectance errors or aerosol type mismatch). The results show that \ourmodel\ is not misled by these potential satellite artifacts. By prioritizing the hard constraints from the ground truth (via $\mathcal{L}_{infer}$) and only using AOD for gradient guidance where appropriate, the model avoids "hallucinating" false hotspots, ensuring the inference remains anchored to reliable ground-level reality.

    \item \textbf{Scenario (d): Magnitude Bias (Calibration).} 
    In this scenario, the AOD field correctly captures the spatial extent of the pollution but significantly overestimates the intensity (indicated by the purple-yellow scale mismatch). \ourmodel\ demonstrates a sophisticated decoupling capability: it extracts the valid \textit{spatial shape} (gradient structure) from the AOD signal while simultaneously calibrating the \textit{absolute magnitude} using the ground stations. This proves that our gradient-based loss formulation effectively filters out systematic biases in satellite retrieval, retaining only the useful structural information.
\end{itemize}

In summary, these cases confirm that \ourmodel\ provides a "best-of-both-worlds" solution. It leverages the fine-grained spatial structure of satellite data when available and consistent, yet seamlessly reverts to robust physical interpolation when satellite data is missing or unreliable.

\section{Discussion}
\label{sec:Discussion}

The development of \ourmodel\ represents a fundamental paradigm shift in urban air quality analytics—transitioning from \textit{point-based temporal forecasting} to \textit{field-based spatiotemporal inference}. By synergizing deep representation learning with atmospheric physical laws, this study \textbf{effectively mitigates} the "blind spot" dilemma that has long plagued environmental monitoring networks.

\subsection{Interpretation of Physical-Data Synergy}
The superior performance of \ourmodel, particularly in the challenging winter season (MAE reduced by 25.2\% vs. SOTA), provides empirical validation for the "Physics-Embedded" design philosophy. 
\begin{itemize}
    \item \textbf{Decoupling Transport Mechanisms:} Unlike traditional GNNs that rely on static geometric distances, \ourmodel\ explicitly disentangles pollutant transport into \textit{isotropic diffusion} and \textit{anisotropic advection}. This allows the model to capture the dominant wind-driven transmission corridors in the BTHSA region—a critical feature for modeling winter haze episodes driven by monsoon winds.
    \item \textbf{Robustness via Structural Constraints:} A critical insight from this work is the advantage of using satellite data as a \textit{loss constraint} rather than an \textit{input feature}. Our analysis (Figure~\ref{fig:grid_inference}) demonstrates that this strategy effectively decouples the model's inference capability from the availability of real-time satellite data. By learning the "spatial gradient patterns" during training, the model internalizes the structural logic of pollution distribution. This enables a \textbf{"physics fallback" mechanism}: when satellite "eyes" are blinded by clouds or darkness, the model seamlessly reverts to physics-driven extrapolation based on meteorology and ground observations, ensuring continuous operation.
\end{itemize}

\subsection{Implications for Urban Environmental Management}

The inductive inference capability of \ourmodel\ \textbf{enables more effective and innovative applications} for smart cities:

\begin{itemize}
    \item \textbf{Virtual Sensing Network:} The model effectively functions as a cost-efficient "virtual sensor," providing reliable air quality estimates for unmonitored suburbs and rural areas. This capability is crucial for identifying hidden pollution hotspots and assessing environmental justice issues where low-income communities often lack monitoring infrastructure.
    \item \textbf{Real-Time Decision Support:} Compared to chemical transport models (CTMs) like CMAQ, which require heavy computational resources and hours of runtime, \ourmodel\ can generate regional inference maps in seconds. This efficiency allows policymakers to perform rapid assessment and emergency response during pollution events.
\end{itemize}

\subsection{Limitations and Future Prospects}
Despite these advancements, challenges remain. 
First, as seen in the extreme case of Station 1057A (Figure~\ref{fig:sparse_neighbors}), the model's ability to capture high-frequency fluctuations degrades in regions with extremely sparse upstream neighbors. While the physical kernels ensure a correct baseline trend, capturing rapid local variations still requires minimal data support. 
Second, the model's reliance on reanalysis meteorology (ERA5) implies that errors in wind field data could propagate into the inference results, highlighting the need for high-precision local meteorological inputs.
Future research could explore: (1) integrating real-time emission proxies (e.g., traffic flow) to enhance local responsiveness; and (2) extending the physics-guided framework to reactive pollutants like Ozone ($O_3$) by incorporating chemical reaction modules.

\section{Conclusion}
\label{sec:conclusion}

To address the practical challenge of \textbf{spatiotemporal inference} in unmonitored urban areas, this study proposes \ourmodel, a physics-guided inductive graph neural network. We successfully reconstruct high-resolution \PM\ fields from sparse ground observations by innovating in three dimensions:

\begin{enumerate}
    \item \textbf{Architecture:} We embed atmospheric fluid dynamics into the deep learning structure via parallel Advection and Diffusion kernels, ensuring the model respects physical transport laws rather than merely fitting statistical correlations.
    \item \textbf{Strategy:} We adopt a dynamic node-masking training strategy that empowers the model with inductive reasoning, enabling generalization to arbitrary unobserved locations.
    \item \textbf{Data Fusion:} We pioneer a \textit{Masked AOD Spatial Gradient Loss}, transforming satellite data from an error-prone input into a robust structural constraint. This novel approach effectively solves the pervasive "missing data" problem caused by cloud cover, achieving all-weather inference capability.
\end{enumerate}

Extensive validation in Beijing-Tianjin-Hebei and Surrounding Areas demonstrates that \ourmodel\ establishes a new state-of-the-art, reducing inference error by over 25\% compared to leading baselines and generating physically plausible pollution maps even under total satellite data loss. This work provides a scalable, low-cost, and robust solution for next-generation environmental sensing, paving the way for data-driven, precision-guided sustainable urban governance.

\section*{Acknowledgments}
\label{sec:Acknowledgments}

We extend our gratitude to the Ministry of Ecology and Environment of the People's Republic of China, the Japan Meteorological Agency (JMA), and the European Centre for Medium-Range Weather Forecasts (ECMWF) for providing the essential data supporting this study. 

This work was supported by the National Natural Science Foundation of China (Grant Nos. 42450183, 12275020, 12135003, 12205025, and 42461144209) and the Ministry of Science and Technology of China (Grant No. 2023YFE0109000). Jingfang Fan is supported by the Fundamental Research Funds for the Central Universities. Shuo Wang acknowledges financial support from the China Scholarship Council (CSC) under Grant No. 202106040117. Mengfan Teng acknowledges support from the Jiangxi University of Science and Technology High-level Talent Project (Grant No. jxust-66) and the Jiangxi Province Youth Science Foundation (Grant No. 20252BAC200051).

\bibliographystyle{elsarticle-harv} 
\bibliography{references}

\end{document}